\documentclass[letterpaper, 10 pt, conference]{ieeeconf}  % Comment this line out if you need a4paper

\IEEEoverridecommandlockouts                              % This command is only needed if 
\usepackage{graphicx}
\usepackage{amsmath} % assumes amsmath package installed

\usepackage[linkcolor=blue,colorlinks=true]{hyperref}
\usepackage{multirow}
\usepackage{gensymb}
\usepackage{float}
\usepackage{amssymb}
\usepackage{subcaption} 
\usepackage[ruled,linesnumbered]{algorithm2e}
\usepackage{algorithmic}
\usepackage[dvipsnames]{xcolor}
\usepackage{blkarray}
\newcommand{\etal}{\textit{et al}.}

\title{\LARGE \bf
A Factor-Graph Approach for Optimization Problems\\with Dynamics Constraints
}

\author{Mandy Xie, Alejandro Escontrela, Frank Dellaert% <-this % stops a space
\thanks{Mandy, Alejandro, and Frank are at the Georgia Institute of Technology}%
\thanks{Atlanta, GA 30332, USA}
\thanks{{\tt(manxie|aescontrela|fd27)@gatech.edu}}%
}

\begin{document}

\maketitle
\thispagestyle{empty}
\pagestyle{empty}

%%%%%%%%%%%%%%%%%%%%%%%%%%%%%%%%%%%%%%%%%%%%%%%%%%%%%%%%%%%%%%%%%%%%%%%%%%%%%%%%
\begin{abstract}

% This paper describes a unified method solving for inverse, forward, and hybrid dynamics problems for robotic manipulators with either open kinematic chains or closed kinematic loops based on factor graphs. Manipulator dynamics is considered to be a well studied problem, and various different algorithms have been developed to solve each type of dynamics problem. However, they are not easily explained in a unified and intuitive way. In this paper, we introduce factor graphs as a unifying graphical language in which not only to solve all types of dynamics problems, but also explain the classical dynamics algorithms in a unified framework.
In this paper, we introduce \emph{dynamics factor graphs} as a graphical framework to solve dynamics problems and kinodynamic motion planning problems with full consideration of whole-body dynamics and contacts. A factor graph representation of dynamics problems provides an insightful visualization of their mathematical structure and can be used in conjunction with sparse nonlinear optimizers to solve challenging, high-dimensional optimization problems in robotics. We can easily formulate kinodynamic motion planning as a trajectory optimization problem with factor graphs. We demonstrate the flexibility and descriptive power of dynamics factor graphs by applying them to control various dynamical systems, ranging from a simple cart pole to a 12-DoF quadrupedal robot.
\end{abstract}

%%%%%%%%%%%%%%%%%%%%%%%%%%%%%%%%%%%%%%%%%%%%%%%%%%%%%%%%%%%%%%%%%%%%%%%%%%%%%%%%
\section{Introduction \& Related Work}
Rigid-body dynamics is a fundamental problem that is often involved in the study of robotics. Specifically, roboticists use inverse dynamics algorithms to compute motor torques in many control problems and forward dynamics algorithms in simulators. Researchers have proposed a variety of algorithms to solve the inverse dynamics problem, including RNEA~\cite{Luh80jdsmc_manipulator,Felis17jar_rbdl, Orin79jmb_NE}. Similarly, algorithms like the Composite-Rigid-Body Algorithm (CRBA)~\cite{Walker82jdsmc_dynamics,Featherstone00icra_crba}, and the Articulated-Body Algorithm (ABA)~\cite{Featherstone83ijrr_aba} have been proposed to solve the forward dynamics problem. Rodriguez~\cite{Rodriguez87jra_forward-inverse-dynamics,Rodriguez86isrm_forward-dynamics} built a unified framework based on the concept of filtering and smoothing to solve both inverse and forward dynamics problems. Based on the work by Rodriguez \etal~\cite{Rodriguez91ijrr_soa}, Jain~\cite{Jain91jgcd_unified, Jain11msd_graph1, Jain11msd_graph2, Jain12ND_multibody1} applied graph theory to dynamics problems, and analyzed various algorithms to solve dynamics problem in a unified formulation. Ascher \etal~\cite{Ascher97ijrr_forward-dynamics} unified the derivation of CRBA and ABA as two elimination methods used to solve forward dynamics. Various physics engines implement these algorithms, including RBDL~\cite{Felis17jar_rbdl}, Bullet Physics~\cite{Coumans19}, ODE~\cite{Smith05ode}, MuJoCo~\cite{Todorov12mujoco}, DART~\cite{Lee18dart}, Pinocchio~\cite{Carpentier-sii19}, etc. However, these dynamics algorithms are not intuitive to understand, and they usually act like black boxes in those physics engines. Hence, it is difficult to leverage these tools to solve various practical problems with dynamics involved, such as optimal control, kinodynamic motion planning, system modeling, state estimation, and simulation.

Developing a single mathematical representation that allows for optimization of controls for arbitrary dynamical systems is not trivial. Much prior work requires expert knowledge of a system's dynamics to develop motion planning algorithms tailored to a particular system. In legged locomotion, various simplified dynamics models have been proposed to enable the control of legged robots. For instance, Kajita \etal~\cite{Kajita01RSJ} model a bipedal robot as a linear inverted pendulum, which enables control of the robot's center of mass position along a plane. Blickhan \etal~\cite{Blickhan89JBiomech} further refine this dynamics model by adding a spring element, thereby enabling more sophisticated trotting and hopping gaits. Dai \etal~\cite{Dai14RAS} employs the \emph{centroidal dynamics} model, which neglects the robot's time-varying inertia matrix and uses Newton's second law of translation and rotation to model the effect of contact forces on the robot's linear and angular acceleration. Researchers studying the control of UAVs also utilize simplified dynamics models \cite{Wang16ICAMechS}. These simplified dynamics models are often exploited in trajectory optimization to optimize over motion plans \cite{Winkler18ral_gait, Winkler2017arXiv}. However, extending these trajectory optimization techniques to control other robotic systems requires significant effort and expert knowledge, as careful consideration of the robot's dynamics is required to ensure that optimized trajectories are still valid for a new system. Posa \etal~ \cite{Posa13springer_TO_dynamical_systems} introduces a framework that allows for trajectory optimization of dynamical systems with contacts but only demonstrates its application to a single planar bipedal robot. Mordatch \etal~\cite{Mordatch12ACM_CIO} introduces \emph{Contact-Invariant Optimization}, a behavior synthesis framework that leverages a simplified dynamics model to produce a wide variety of human behaviors for animated characters.

In this work, we introduce dynamics factor graphs (DFGs) as a framework for solving classical dynamics problems and advanced problems in control and motion planning. We leverage the descriptive power of DFGs to model a variety of systems, ranging from the nonlinear cart-pole to a 7-DoF Kuka Arm to a 12-DoF quadrupedal robot. We then incorporate DFGs into a trajectory optimization framework and use state-of-the-art sparse nonlinear optimizers to generate motion plans for various robots, all without requiring simplified dynamics models or expert knowledge of the robot's dynamics a-priori. The main contributions are:
\begin{itemize}
\item A graphical representation of dynamics problems with factor graphs;
\item Demonstration of solving a large variety of problems in robotics with factor graphs, including classical dynamics, dynamics with constraints and objectives, and kinodynamic motion planning;
\item Demonstration of using a state-of-the-art sparse nonlinear optimizer based on GTSAM and a hinge loss formulation of inequality constraints to solve high-dimensional optimization problems in robotics.
% \item The graphical framework for dynamics
% \item We show how a large variety of optimization problems with dynamics constrained can be solved in this framework, including...

% \item We show how to use a sota sparse nonlinear optimizer based on GTSAM and a hinge loss formulation of inequality constraints enables us to solve all of the above problems. (Later in the paper, mention we follow CHOMP etc...). Go into further detail regarding sparsity, GTSAM later in the paper... Say: "hinge loss... which we have found to lead to large convergence spaces".

% \item a graphical representation for dynamics problems based on factor graphs, which provides an insightful visualization of the underlying mathematics;
% \item application of our dynamics factor graph to solve inverse, forward, and hybrid dynamics problems with added support for constraints and objectives.
% \item a kinodynamic motion planning framework that applies dynamics factor graphs to solve trajectory optimization problems for arbitrary dynamical systems.
% \item a factor graph which can solve inverse, forward and hybrid dynamics problems;
% \item a unified method which can solve inverse, forward and hybrid dynamics problems;
\end{itemize}

\section{Review of Manipulator Dynamics}
We briefly review the modern geometric view of robot dynamics and follow the exposition and notation from the recent text by Lynch and Park~\cite{Lynch17book_robotics}. As convincingly argued in their introduction, this geometric view pioneered by Brockett~\cite{Brockett84mtns_robotics} and Murray \etal~\cite{Murray94book} unlocks the powerful tools of modern differential geometry to reason about robot dynamics. It will also help below in describing the various dynamics algorithms in a concise graphical representation.
 
% macros for traditional NE
\newcommand{\I}{\mathcal{I}}
\newcommand{\w}{\omega}
\newcommand{\wdot}{\dot{\w}}

% Traditionally, the Newton-Euler equations of motion for a rigid body moving in space subjects to external forces $f$ and torques $\tau$, can be expressed in body coordinates as (Equations 8.22 and 8.23 on page 242 in \cite{Lynch17book_robotics}),
% \begin{align} 
% \label{eq:rtd}
% f_b &= m\dot{v}_b + \w_b \times mv_b \\
% \label{eq:rrd}
% \tau_b &= \I_b \wdot_b + \w_b \times \I_b \w_b
% \end{align}
% with $m$, $\I_b$, $v_b$, and $\w_b$ respectively the mass, inertia, linear and angular velocity expressed in body coordinate frame.

% macros for link and joint quantities
\newcommand{\V}{\mathcal{V}}
\newcommand{\Vdot}{\mathcal{\dot{V}}}
\newcommand{\G}{\mathcal{G}}
\newcommand{\F}{\mathcal{F}}
\newcommand{\q}{\theta}
\newcommand{\qdot}{\dot{\q}}
\newcommand{\qddot}{\ddot{\q}}
\newcommand{\Axis}{\mathcal{A}}

% In the geometric view, we combine equations \eqref{eq:rtd} and \eqref{eq:rrd} to obtain an equation in terms of the six-dimensional body wrench $\F_b$ and body twist $\V_b$ quantities (Equation 8.40 on page 247 of \cite{Lynch17book_robotics}),
% \begin{equation} \label{eq:crd}
%     \F_b = \G_b \dot{\V}_b - [ad_{\V_b}]^T \G_b \V_b
% \end{equation}
% where the new quantities are defined as
% $$
% \V_b =  \left[ \begin{array}{c} \w_b \\ v_b \end{array} \right] \ \
% \F_b =  \left[ \begin{array}{c} \tau_b \\ f_b \end{array} \right] $$
% $$\G_b = \begin{bmatrix} \I_b & 0 \\ 0 & m I \end{bmatrix} \ \
% [ad_{\V_b}] = \begin{bmatrix} [\w_b] & 0 \\ [v_b] & [\w_b] \end{bmatrix}
% $$
% Above $[\w_b]$ is the skew-symmetric matrix formed from $\w_b$, i.e.,
% $$
% [\w_b]=R^T \dot{R}
% $$
% with $R\in SO(3)$ the rotation associated with a link.

Closely following Section 8.3 in~\cite{Lynch17book_robotics}, applying this to the links of a serial manipulator and taking into account the constraints at the joints, we obtain four equations relating both link and joint quantities.
In particular, the twist and acceleration $\V_i$ and $\Vdot_i$ for the $i$\textsuperscript{th} link are expressed in a body-fixed coordinate frame rigidly attached to the link. The wrench transmitted through joint $i$ is denoted as $\F_i$, and $\G_i$ is the link's inertia matrix. Without loss of generality, below we assume all rotational joints, and we then have: 
\begin{align} 
\label{eq:twist}
    \V_i - [Ad_{T_{i,i-1}(\q_i)}]\V_{i-1} - \Axis_i\qdot_i &= 0 \\
\label{eq:accel}
    \Vdot_i - [Ad_{T_{i,i-1}(\q_i)}]\Vdot_{i-1} - \Axis_i\qddot_i - [ad_{\V_i}]\Axis_i\qdot_i &= 0  \\
\label{eq:wrench}
    Ad^T_{T_{i+1,i}(\q_{i+1})}\F_{i+1} - \F_i + \G_i\dot{\V}_i - [ad_{\V_i}]^T\G_i\V_i &= 0 \\
\label{eq:torque}
    \F_i^T\Axis_i - \tau_i &= 0
\end{align}
where $\Axis_i$ is the screw axis for joint $i$ (expressed in link $i$), and $Ad_{T_{i,i-1}(\q_i)}$ is the adjoint transformation associated with the transform $T_{i,i-1}$ between the links (a function of $\q_i$).
% $$[Ad_{T_{i,i-1}(\q_i)}] = \begin{bmatrix} R_{i,i-1} & 0 \\ [p]_\times R_{i, i-1} & R_{i, i-1}\end{bmatrix}$$

The four equations \eqref{eq:twist}-\eqref{eq:torque} express the dynamic constraints between link $i$ and link $i-1$ imposed by joint $i$: 
\eqref{eq:twist} describes the relationship between twist $\V_i$ and twist $\V_{i-1}$, where $\qdot_i$ is the angular velocity of joint $i$; 
\eqref{eq:accel} describes the constraint between acceleration $\Vdot_i$ and acceleration $\Vdot_{i+1}$, which involves components due to joint acceleration $\qddot_i$ and the acceleration caused by rotation;
\eqref{eq:wrench} describes the balance between the wrench $\F_i$ through joint $i$ and the wrench $\F_{i+1}$ applied through joint $i+1$;
\eqref{eq:torque} describes that the torque applied at joint {i} equals to the projection of wrench $\F_i$ on the screw axis $\Axis_i$ corresponding to joint $i$. Gravity is not considered above but can easily be accounted for using a "trick" described by Lynch \& Park~\cite{Lynch17book_robotics} that adds an extra acceleration term to the base. 

% While clever, we have found it more intuitive to explicitly deal with gravity in our implementation, where we simply add a gravity term to the wrench equation \eqref{eq:wrench}, expressed in each link's body frame.
% \input{sub_tex/2-manipulator_dynamics_review.tex}

\section{A Factor Graph Approach}
This paper proposes to represent optimal control problems involving dynamic constraints using a factor graph~\cite{Kschischang01it}, a graphical model to describe the structure of sparse computational problems. A factor graph consists of \emph{factors} and \emph{variables}, where factors correspond to objectives, equality constraints, and inequality constraints involving the variables being optimized over. Variables are only connected to factors that they are involved in, and the resulting bipartite graph reveals the sparsity of the computation and enables the use of efficient optimization techniques. Factor graphs have been used in constraint satisfaction~\cite{Seidel81ijcai,Freuder82jacm,Dechter87ai}, AI~\cite{Pearl88book,Shenoy86expert,Lauritzen88jrssb, Frey97ccc,Kschischang01it}, sparse linear algebra~\cite{George84laa,Gilbert93chapter,Heggernes96siam}, information theory~\cite{Tanner81it,Loeliger04spm}, combinatorial optimization~\cite{Bertele72book,Bertele72jmaa,Bertele73jct}, and query theory~\cite{Beeri81stoc,Goodman82tods}. They have been successfully applied in other areas of robotics, such as SLAM~\cite{Kaess08tro,Kaess12ijrr,Dellaert17fnt_fg,Forster17tro} and state estimation in legged robots~\cite{Hartley18icra_fg_estimation, Wisth19ral_legged_fg}.
 
\begin{figure}[!t]%[!htb]
 	\vspace{1mm}
	\centering
	\includegraphics[scale=0.85]{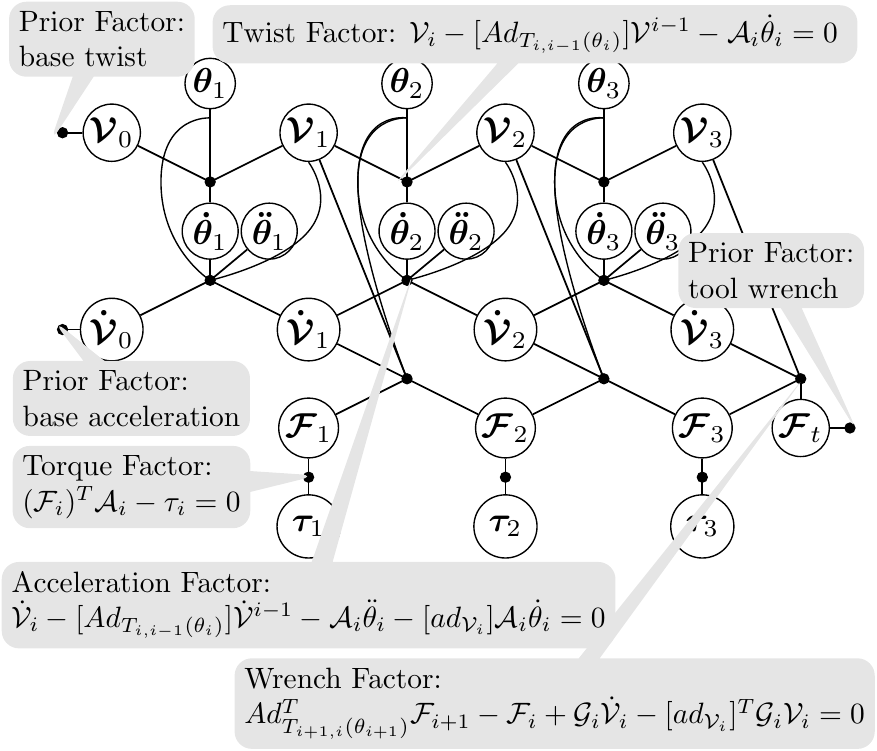}
	\caption{Dynamics factor graph for a RRR robot.}
	\label{fig:DFG-rrr}
	\vspace{-4mm}
\end{figure}
 
\subsection{Factor Graphs for Constrained Optimization}
A constrained optimization problem may be written as
\begin{equation}
\begin{aligned}
\min_{\boldsymbol{\chi}} \quad & f(\boldsymbol{\chi})\\
\textrm{s.t.} \quad & g_i(\boldsymbol{\chi}) = 0 \, \text{ for } i=1,\dots, n \\
                    & h_j(\boldsymbol{\chi}) \leq 0 \text{ for } j=1,\dots, m \\
\end{aligned}
\end{equation}
where $g_i(\boldsymbol{x})$, $i=1,\dots,n$ and $h_j(\boldsymbol{x})$, $j=1,\dots,m$ are equality and inequality constraints, respectively and $f(\boldsymbol{x})$ is an objective function which we wish to optimize subject to the constraints. We convert the constrained problem into an unconstrained problem as follows:
\begin{equation}
\begin{aligned}
\boldsymbol{\chi}^* =& \min_{\boldsymbol{\chi}} \|r_f(\boldsymbol{\chi})\|^2_{\Sigma_f} + \sum_{i=1\dots\text{n}}\|r_{g,i}(\boldsymbol{\chi})\|^2_{\Sigma_{g, i}}\\
&\qquad\qquad\qquad\,\,\, + \sum_{j=1\dots\text{m}}\|r_{h,j}(\boldsymbol{\chi})\|^2_{\Sigma_{\tilde{h}, j}}
\end{aligned}
\end{equation}
where $r_f(\boldsymbol{\chi}) = \log{f(\boldsymbol{\chi})}$ is the log value of the objective function, $r_{g, i}(\boldsymbol{\chi}) = \log{g_i(\boldsymbol{\chi})}$ are the log residual error functions associated with equality constraints, and $r_{h, j}(\boldsymbol{\chi}) = \log{\tilde{h}_j(\boldsymbol{\chi})} = \log{\max(0, h_j(\boldsymbol{\chi}))}$ is the log of a hinge loss approximation of the residual error associated with the inequality constraints. The objective function $f(\boldsymbol{\chi})$, equality constraints $g_i(\boldsymbol{\chi})$, and inequality constraints $h_j(\boldsymbol{\chi})$ may all be represented as \emph{factors} in a factor graph. Fig.~\ref{fig:DFG-rrr-ad} shows a schematic representation of a factor graph, where solid nodes correspond to objective functions or constraints.
% In the following section, we explain how to model the dynamics constraints as a set of equality constraint factors. Later, in section \ref{sec:solving_constrained_dynamics_problems}, we demonstrate how to incorporate inequality constraints and objective functions into the factor graph.

% A general theory specified in terms of algebraic semirings was also developed~\cite{Carre71jima}, and seminal work in theory proved essential computational complexity results~\cite{Lipton79nd} based on
% the existence of separator theorems for certain classes of graphs
% ~\cite{Lipton79sep}.

\subsection{Dynamics Factor Graphs (DFGs)}

% The following describes how to use factor graphs in solving manipulator dynamics problems, and benefit from its advantage. 
We use a factor graph to represent the structure of the dynamics constraints \eqref{eq:twist}-\eqref{eq:torque} for a particular robot configuration. Fig.~\ref{fig:DFG-rrr} illustrates this for a serial chain comprised of three revolute joints (RRR). Variables including twists $\V_i$, accelerations $\Vdot_i$, wrenches $\F_i$, joint angles $\q_i$, joint velocities $\qdot_i$, joint accelerations $\qddot_i$, and torques $\tau_i$ are constrained to satisfy the rigid-body dynamics equations \eqref{eq:twist}-\eqref{eq:torque}. We can use the dynamics factor graph corresponding to all variables and constraints to model various dynamical systems, ranging from serial manipulators with fixed bases to legged robots with floating bases. We can also use them to solve classical dynamics problems (i.e., inverse and forward dynamics problems) and optimization problems in motion planning, as discussed in the sections below.

\section{Solving Classical Dynamics Problems}\label{sec:solving_classical_dynamics_problems}
This section demonstrates how to solve inverse and forward dynamics problems with DFGs and illustrate the classical dynamics algorithms within this framework.
\subsection{Inverse Dynamics}
\begin{figure}%[!htb]
    \vspace{1mm}
	\centering
	\begin{subfigure}[b]{0.2\textwidth}
		\centering
		\includegraphics[scale=0.8]{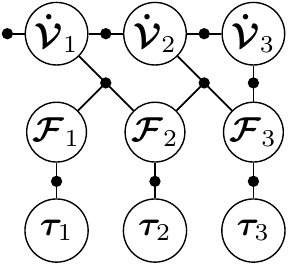}
		\caption{Simplified inverse DFG}
		\label{fig:IDFG-linear}
	\end{subfigure}
% 	\begin{subfigure}[b]{0.2\textwidth}
% 		\centering
% 		\includegraphics[scale=0.15]{figures/inverse_sparse_3R.png}
% 		\caption{}
% 		\label{fig:ID-sparse}
% 	\end{subfigure}
% 	\caption{(a) Simplified inverse dynamics factor graph (b) Block-sparse matrix corresponding to Fig.~\ref{fig:IDFG-linear}}
% 	\label{fig:IDFG-sparse}
	\centering
	\begin{subfigure}[b]{0.2\textwidth}
		\centering
		\includegraphics[scale=0.8]{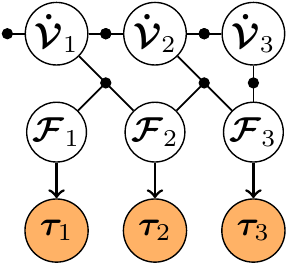}
		\caption{Eliminate all the $\tau$.}
		\label{fig:elimination1}
	\end{subfigure}
	\begin{subfigure}[b]{0.2\textwidth}
		\centering
		\includegraphics[scale=0.8]{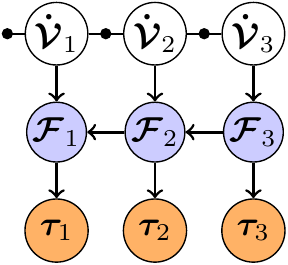}
		\caption{Eliminate all the $\F$.}
		\label{fig:elimination3}
	\end{subfigure}
	\begin{subfigure}[b]{0.2\textwidth}
		\centering
		\includegraphics[scale=0.8]{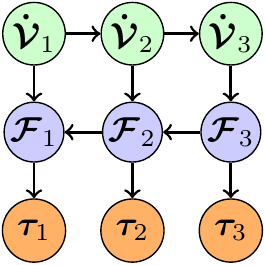}
		\caption{Eliminate all the $\Vdot$.}
		\label{fig:elimination4}
	\end{subfigure}
	\caption{Steps in elimination on the inverse DFG.}
	\label{fig:IDFG-elimination}
	\vspace{-4mm}
\end{figure}

In inverse dynamics, we seek the joint torques $\tau_i$ to realize the desired joint accelerations $\qddot_i$. We can construct a simplified, less cluttered, \emph{inverse dynamics graph} by replacing all known variables with constants in the factors to which they are connected. Fig.~\ref{fig:IDFG-linear} shows the DFG for a RRR robot, corresponding to the nine linear constraints of the 3R inverse dynamics problem. Solving the inverse dynamics problem is equivalent to solving this factor graph, and it can be done by back-substitution after performing elimination on the graph.

We illustrate the elimination process in the 3R case for a particular ordering in Fig.~\ref{fig:IDFG-elimination}. 
The elimination is performed in the order $\tau_3 \ \tau_2 \ \tau_1$, $\F_1 \ \F_1 \ \F_3$, $\Vdot_3 \ \Vdot_3 \ \Vdot_1$.
% The elimination is performed in the order shown in Tab.~\ref{ID-EO}.
Fig.~\ref{fig:elimination1} shows the result of first eliminating the torques $\tau_i$, where the arrows show that the torques $\tau_i$ only depend on the corresponding wrenches $\F_i$.
We then eliminate $\F_1$, which results in a dependence of $\F_1$ on $\Vdot_1$ and $\F_2$. 
After eliminating all the wrenches, we get the result as shown in Fig.~\ref{fig:elimination3}.
Finally, we eliminate the all the twist accelerations $\Vdot_3 \ \Vdot_3 \ \Vdot_1$, in that order. After completing all these elimination steps, the inverse dynamics factor graph in Fig.~\ref{fig:IDFG-linear} is thereby converted to the directed acyclic graph (DAG) as shown in Fig.~\ref{fig:elimination4}.

After elimination, back-substitution \emph{in reverse elimination order} solves for the values of all intermediate quantities and the desired torques. For the example ordering in \ref{fig:IDFG-elimination}, back-substitution first computes the 6-dimensional accelerations $\Vdot_i$, then the link wrenches $\F_i$, and finally the scalar torques. This \emph{exactly} matches the forward-backward path used by the recursive Newton-Euler algorithm (RNEA)~\cite{Luh80jdsmc_manipulator}. 

% Therefore, the resulting DAG is a graphical representation of RNEA, where the green and blue/orange colors resp. correspond to the forward path and the backward path.

\subsection{Forward Dynamics}
\begin{figure}%[!htb]
    \vspace{1mm}
	\centering
	\begin{subfigure}[b]{0.15\textwidth}
		\centering
		\includegraphics[scale=0.8]{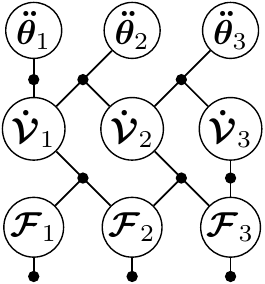}
		\caption{}
		\label{fig:FDFG-linear}
	\end{subfigure}
	\begin{subfigure}[b]{0.15\textwidth}
		\centering
		\includegraphics[scale=0.8]{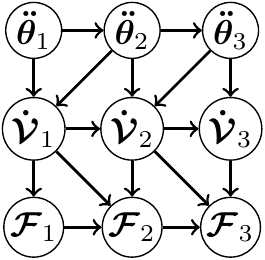}
		\caption{}
		\label{fig:FDBN-CRBA}
	\end{subfigure}
	\begin{subfigure}[b]{0.15\textwidth}
		\centering
		\includegraphics[scale=0.8]{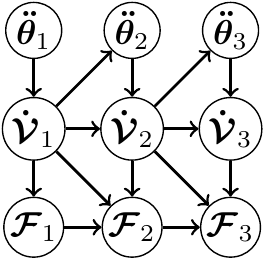}
		\caption{}
		\label{fig:FDBN-ABA}
	\end{subfigure}
	\caption{(a) Simplified forward DFG; (b) CRBA ordering DAG; (c) ABA ordering DAG.}
	\vspace{-4mm}
\end{figure}

In traditional expositions, the forward dynamics problem cannot be solved as neatly as the relatively easy inverse dynamics problem.
In the forward case, we seek the joint accelerations $\qddot$ when given $\q$, $\qdot$, and $\tau$. Similar to the inverse dynamics factor graph, we can simplify the forward dynamics factor graph, as shown in Fig.~\ref{fig:FDFG-linear}. 

In the same spirit as our work, Ascher \etal~\cite{Ascher97ijrr_forward-dynamics} shows that two of the most widely used forward dynamics algorithms, CRBA~\cite{Featherstone00icra_crba} and ABA~\cite{Featherstone83ijrr_aba}, can be explained as two different elimination methods to solve the same linear system. 

In our framework, CRBA and ABA can also be visualized as two different DAGs resulting from solving the forward dynamics factor graph with two different elimination orders shown in Fig.~\ref{fig:FDBN-CRBA} and Fig.~\ref{fig:FDBN-ABA}.
CRBA first eliminates all the wrenches $\F_i$, then eliminates all the accelerations $\Vdot_i$, and eliminates all the angular accelerations $\qddot_i$ last.
In contrast, in ABA we alternate between eliminating the wrenches $\F_i$, accelerations $\Vdot_i$ and angular accelerations $\qddot_i$ for $i\in n\dots1$.
We can view the resulting DAGs as graphical representations of CRBA and ABA, and for a given robot configuration, we can write a custom back-substitution program in reverse elimination order.

\section{Solving Constrained Dynamics Problems} \label{sec:solving_constrained_dynamics_problems}
\begin{figure}[!t]%[!htb]
    \vspace{1mm}
	\centering
	\includegraphics[scale=0.85]{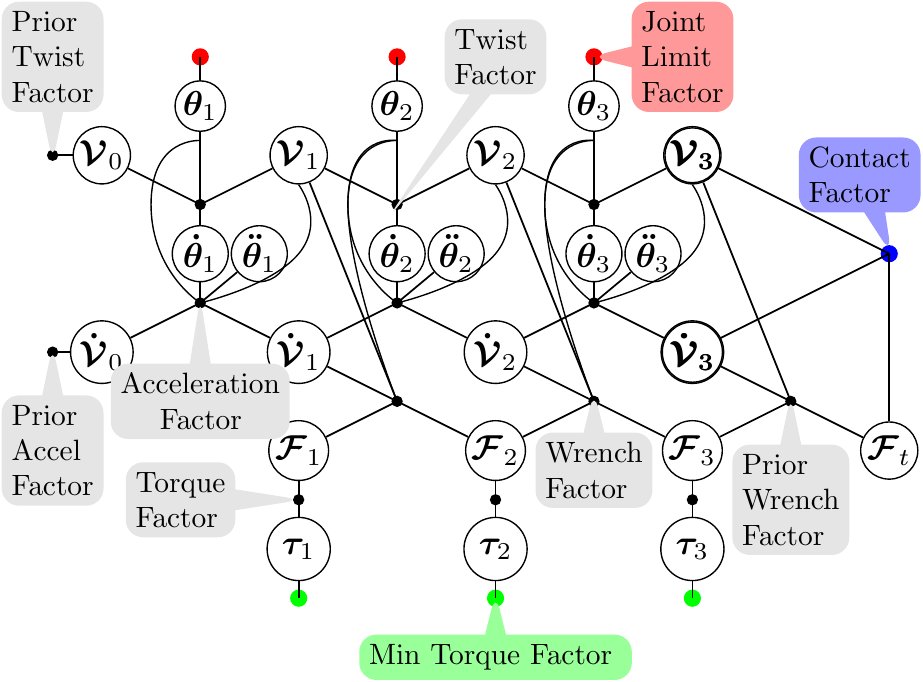}
	\caption{Constrained dynamics factor graph for a RRR robot.}
	\label{fig:DFG-rrr-ad}
	\vspace{-4mm}
\end{figure}
This section briefly explains how to solve dynamics problems with joint limit constraints, contacts,  task-dependent objectives, and redundancy resolution. The complexity of practical problems in robotics far surpasses that of the previous section's classical problems. For instance, the system has to stay within its joint limits at any time instance to avoid damage. Redundancy issues often occur when the system is not fully constrained, for instance, solving inverse dynamics for a manipulator with kinematic loops or planning motion for tasks in which we constrain the external wrench acting on the end-effector in only one direction. Also, when there is contact involved in the motion, the dynamics problem becomes even more complicated. In motion planning or optimal control literature, joint limits and contacts are usually represented as constraints in the optimization problem, and the redundancy issue can be resolved by optimizing for a minimum torque objective.

Here, we propose to solve all these problems with DFGs by incorporating them as factors in the graph, as shown in Fig.~\ref{fig:DFG-rrr-ad}. We then solve the constrained optimization dynamics problem with GTSAM~\cite{Dellaert12report_gtsam,Dellaert17fnt_fg}.

\subsection{Joint Limit Factors}
Joint limits are formulated as the following inequality constraint for each joint angle $q_i$,
$h_{\text{lim}}(q_i) - \theta_\text{lim, i} \leq 0$,
where $\theta_\text{lim, i}$ is the limit for the i\textsuperscript{th} joint. We incorporate this inequality in a hinge loss function:
\begin{equation}\label{eq:hinge}
\tilde{h}_\text{lim}(q_i)=\left\{
\begin{array}{ll}
\alpha(\theta_\text{lim, lower} - q_\text{i} + \varepsilon) &{\text{if } \ q_\text{i} - \theta_\text{lim, i} \leq \varepsilon}\\
\alpha(q_\text{i} -  \theta_\text{lim, upper} + \varepsilon) &{\text{if } \ \theta_\text{lim, upper} - q_\text{i} \leq \varepsilon}\\
0\ &{\text{otherwise}}
\end{array}\right.
\end{equation}
where $\theta_\text{lim, lower}$ is the lower limit, $\theta_\text{lim, upper}$ is the upper limit, $\varepsilon$ is a constant threshold to prevent exceeding the limit, and $\alpha$ is a constant ratio which determines how fast the error grows as the value approaches the limit. If the value is within the threshold, then the cost is 0. Hence, the limit violations are prevented during the optimization. This technique is characteristic of interior point method~\cite{Dikin67ams_interior_point_methods}. 

% $f(q_i) \leq e_i$, and the cost function $h_l(q_i)$ under the current state can be written as
% \begin{equation}\label{eq:inequality}
% h_l(q_i) = c_l(f(q_i))
% \end{equation}
% where $c_{l}()$ is defined as the hinge loss:
% \begin{equation}\label{eq:hinge}
% c_l(z)=\left\{
% \begin{array}{ll}
% a(z_l - z + \varepsilon) &{if \ z - z_l \leq \varepsilon}\\
% a(z - z_u + \varepsilon) &{if \ z_u - z \leq \varepsilon}\\
% 0\ &{o/w}
% \end{array}\right.
% \end{equation}

\subsection{Minimum Torque Factor} \label{ssec:minimum_torque_factor}
Minimum torque objective is formulated as an equality constraint, and cost function can be expressed as
$g_{\tau}(\tau_i) = \tau_i$,
where $\tau_i$ is the torque at joint $i$. The cost grows as the torque increases, and we can expect that the optimizer solves for a solution with minimum torque values.

\subsection{Contact Factor (Friction Cone)} \label{ssec:contact_factor}
A desirable property of legged locomotion controllers is to provide robustness against slipping motions. This property benefits the robot's stability and improves locomotion efficiency. A common approach to avoid slipping motions is to constrain the contact forces at the end-effectors to lie within a friction cone \cite{Winkler18ral_gait, Grandia19arXiv_feedback_mpc}. This inequality constraint prevents the lateral forces from dominating the resistive Coulomb friction forces, thereby preventing slipping. Consider the external wrench at the robot's i\textsuperscript{th} end-effector $\mathcal{F}_i^e = [m_i^e; f_i^e] \in \chi$. The contact factor enforces the following inequality constraint:
$$\|f_i^e - (f_i^e \cdot n_i)n_i\|_2 \leq \mu_i(f_i^e \cdot n_i)$$

$\mu_i$ is the static friction coefficient at the i\textsuperscript{th} contact, and $b_i$ is the vector normal to the surface. This inequality constraint is enforced using a hinge loss function and incorporated as a factor connected to each end-effector in the DFG.

\section{Motion Planning with Dynamics Factor Graph (DFGP)}
\begin{figure}[!ht]
	\centering
	\includegraphics[scale=1.]{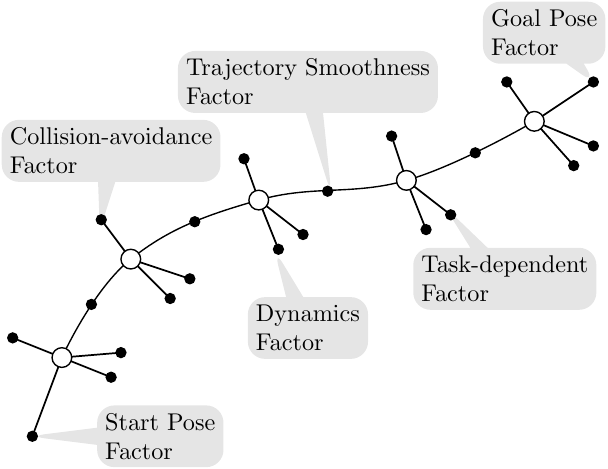}
	\caption{kinodynamic motion planning factor graph, where solid nodes indicate factors and hollow nodes correspond to the decision variables.
	\vspace{-3mm}
% 	To simplify the notation, we use a dynamics factor to represent all the dynamic constraints at each time slice in the factor graph.
	}
	\label{fig:dfgp}
\end{figure}
After solving the constrained dynamics problem, motion planning with the whole-body dynamics constraints becomes straightforward. A simple intuition is that we can view the constrained dynamics problem as a single time instance of the motion planning problem and incorporate the constrained dynamics factor graphs into a trajectory optimization problem in the style of GPMP2~\cite{Mukadam18ijrr_gpmp2}, as shown in Fig.~\ref{fig:dfgp}. To accomplish a real motion planning task, we need to add a few more factors to the graph. For instance, we use initial and goal pose factors to satisfy the demand of moving the end-effector from the initial pose to a goal pose, trajectory smoothness factors are applied to encourage smooth trajectories, and we add collision avoidance factors to ensure collision-free motion. Other task-dependent factors can be easily incorporated into a DFG as well. Due to the page limit, we only briefly describe the trajectory smoothness factor, which is only slightly different from the one outlined in GPMP2. For details on start and goal factors and collision avoidance factors, one can refer to GPMP2~\cite{Mukadam18ijrr_gpmp2}.  

% \subsection{Fixed Start and Goal Poses}
% The cost of start and goal pose constraints can be expressed as a function of the state vectors $x_s$ and $x_g$:
% \begin{align}
% h_s(x_s) &= f(x_s) - p_s \label{eq:startpose}
% \\
% h_g(x_g) &= f(x_g) - p_g \label{eq:goalpose}
% \end{align}
% where $f(*)$ is the forward kinematics which maps any configuration to a workspace (this definition applies to the rest of the paper), $p_s$ is the desired start pose and $p_g$ is the desired goal pose of the end effector. 

\begin{figure}[!b]
	\centering
	\begin{subfigure}[b]{1.0\linewidth}
		\includegraphics[width=0.82\linewidth]{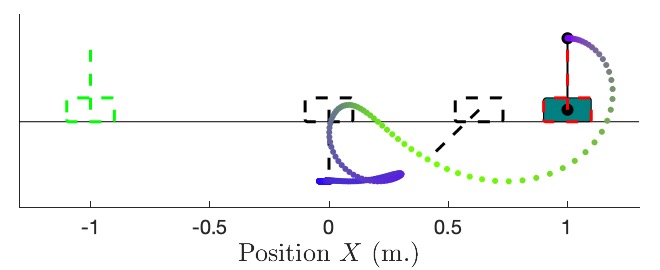}
		\caption{}
		\label{fig:cart_pole_task_0}
	\end{subfigure}
	\begin{subfigure}[b]{1.0\linewidth}
		\includegraphics[width=0.82\linewidth]{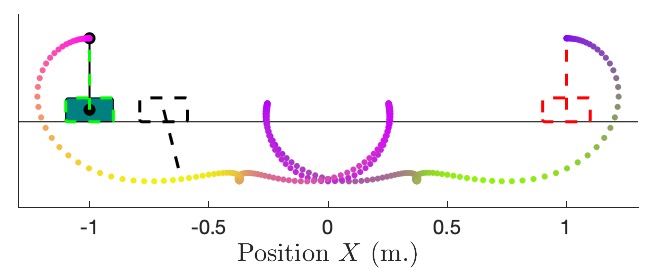}
		\caption{}
		\label{fig:cart_pole_task_1}
	\end{subfigure}
	\begin{subfigure}[b]{1.0\linewidth}
		\includegraphics[width=0.85\linewidth]{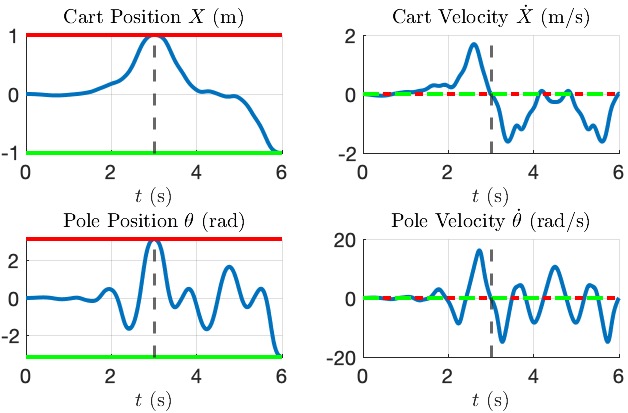}
		\caption{}
		\label{fig:cart_pole_task_2}
	\end{subfigure}
	\caption{The DFGP was used optimize a trajectory for the cart-pole system that achieves two configurations in quick succession. (a) and (b) show the trajectory generated to reach the 1st and 2\textsuperscript{nd} configuration, respectively. (c) demonstrates that the optimized trajectory satisfies the constraints imposed for the 1st (t = 3s) and 2nd configuration (t = 6s), shown as red and green bars, respectively. (See supplemental videos)}
	\label{fig:cart_pole_task}
\end{figure}

\subsection{Trajectory Smoothness Factor} \label{ssec:trajectory_smoothness_factor}
Here we describe a method which we use to optimize smooth trajectories using DFGs and additional trajectory factors. We use a method similar to the one discussed in GPMP2, but with the assumption of a continuous-time configuration space trajectory with a constant acceleration instead of constant velocity, and add a Gaussian process (GP) smoothness prior with cost function,
$g_m(x_{i-1}, x_i) = x_i - \Phi(t_i,t_{i-1})x_{i-1}$
and covariance matrix 
\begin{equation}\label{eq:sigma_accel}
\Sigma = \begin{bmatrix} \frac{1}{2} \Delta t_i^5 \mathbf{Q}_C &
\frac{1}{8} \Delta t_i^4 \mathbf{Q}_C & \frac{1}{6} \Delta t_i^3 \mathbf{Q}_C\\ \frac{1}{8} \Delta t_i^4 \mathbf{Q}_C & 
\frac{1}{3} \Delta t_i^3 \mathbf{Q}_C & \frac{1}{2} \Delta t_i^2 \mathbf{Q}_C \\ \frac{1}{6} \Delta t_i^3 \mathbf{Q}_C & \frac{1}{2} \Delta t_i^2 \mathbf{Q}_C &
\Delta t_i \mathbf{Q}_C \end{bmatrix}
\end{equation}
where $\mathbf{Q}_C$ is the power-spectral density matrix associated with the GP. We also define the state transition matrix 
\begin{equation}\label{eq:phi_accel}
\Phi(t_i,t_{i-1}) = \begin{bmatrix} 1 & \Delta t_i & \frac{1}{2}(\Delta t_i)^2 \\ 0 & 1 & \Delta t_i \\ 0 & 0 & 1 \end{bmatrix}
\end{equation}
associated with a constant acceleration assumption between times $t_{i-1}$ and $t_i$, and $\Delta t_i = t_i - t_{i-1}.$

\subsection{Kinodynamic Motion Planning Factor Graph}
A \emph{kinodynamic motion planning factor graph}, as shown in Fig.~\ref{fig:dfgp}, is designed to solve the kinodynamic motion planning problem\cite{Donald93jacm_KDMP}, in which both constraints and objectives are represented as factors. In particular, we use constrained DFGs for each time instance, and connect them with trajectory smoothness factors (Section \ref{ssec:trajectory_smoothness_factor}) to encourage smooth trajectories. In addition, the initial pose constraint, goal pose constraint, and obstacle avoidance objective are included to fulfill the task while avoiding collisions.

% trajectory smoothness factors (Section \ref{ssec:trajectory_smoothness_factor}) to connect constrained DFGs corresponding to each pair of successive time steps. Constraints include joint limit constraints, initial value constraints, contact constraints, and other task-dependent kinodynamic constraints. Objectives can be used to drive the system to a goal pose, minimize energy expenditure, and ensure smooth and collision-free trajectories. Figure~\ref{fig:dfgp} illustrates the bipartite kinodynamic motion planning factor graph, where solid nodes indicate factors and hollow nodes correspond to the decision variables.

\section{Experiments} \label{sec:experiments}
We implemented DFGP using the GTSAM Factor Graph optimization library~\cite{Dellaert12report_gtsam,Dellaert17fnt_fg}, and ran simulations in V-REP \cite{coppeliaSim} and PyBullet \cite{Coumans19} for visualization. 
% We used the Levenberg-Marquardt trust-region optimization algorithm to solve the nonlinear least squares optimization problem for DFGP, with an initial value for $\lambda = 0.01$, and terminating the optimization process if either of the following conditions is satisfied: 1) it reaches a maximum of 200 iterations, or 2) the relative decrease in error is smaller than $10^{-5}$. 
\subsection{Cart Pole}
The cart-pole system is composed of an unactuated simple pendulum attached to a wheeled cart. A common task in optimal control is to balance the pendulum around its unstable equilibrium, using only horizontal forces on the cart. In this experiment, we tackle a more sophisticated control problem that involves driving the cart-pole system to \emph{multiple goal configurations in quick succession}. We apply our DFGP algorithm to optimize for a single trajectory that enables the cart-pole system to achieve two goal configurations at prescribed times. We apply a zero-torque constraint on the unactuated pendulum. We also apply the trajectory smoothness factors discussed in section \ref{ssec:trajectory_smoothness_factor} and the minimum torque objective discussed in section \ref{ssec:minimum_torque_factor}. Two goal configurations are imposed as objectives in the trajectory factor graph:
\begin{equation}\label{eq:smoothness}
f_{\mathrm{goal, } t= \eta s}(\chi) = \|\theta_{t= \eta s} - \theta_{t= \eta s}^*\| + \|p_{t= \eta s} - p_{t= \eta s}^*\|
\end{equation}
Where $p$ and $\theta$ are the cart-pole's horizontal position and pole angle, respectively. The first configuration requires that the cart-pole come to rest at $[p_\mathrm{t=3s}^*, \theta_\mathrm{t=3s}^*] = [1 m, \pi \mathrm{rad}]$. The second configuration requires that the cart-pole come to rest at $[p_\mathrm{t=6 s}^*, \theta_\mathrm{t=6s}^*] = [-1 m, -\pi\mathrm{rad}]$. Figures~\ref{fig:cart_pole_task_0} and \ref{fig:cart_pole_task_1} visualize the cart-pole's execution of the optimized trajectory. Fig.~\ref{fig:cart_pole_task_2} demonstrates that the DFGP trajectory satisfies both the position and velocity objectives for both configurations.

\subsection{Kuka Arm}
\label{subsec:kuka_arm}
\begin{figure}[!t]
    \vspace{1mm}
	\centering
	\begin{subfigure}[b]{0.23\textwidth}
		\centering
		\includegraphics[scale=0.21]{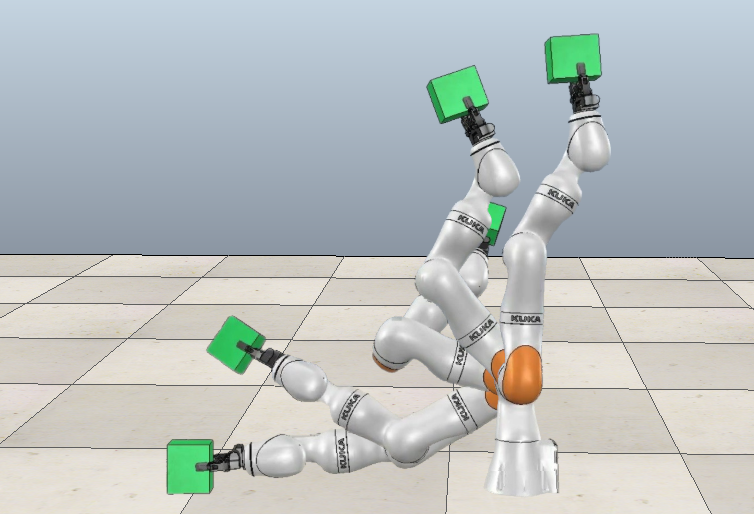}
		\caption{}
		\label{fig:kukaMinTorque}
	\end{subfigure}
	\begin{subfigure}[b]{0.23\textwidth}
		\centering
		\includegraphics[scale=0.21]{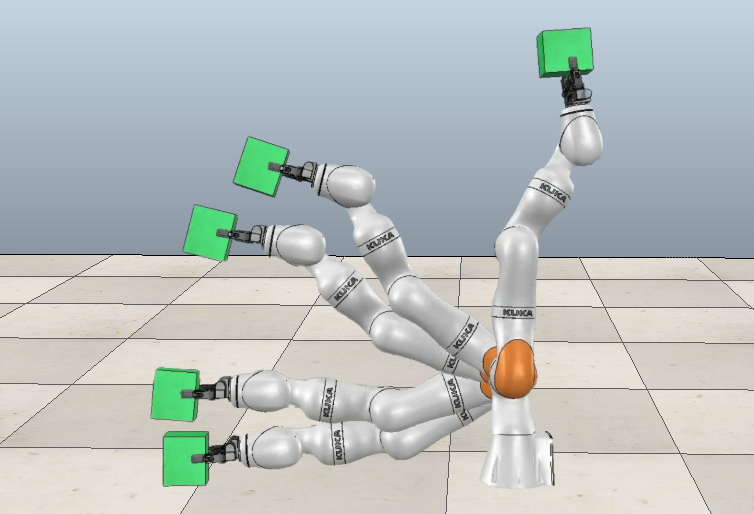}
		\caption{}
		\label{fig:kukaNoMinTorque}
	\end{subfigure}
% 	\caption{(a) shows a solution obtained by DFGP for kinodynamic motion planning, where the objective is to find a trajectory for the KUKA robot to bring a block from the the floor to its upright position. (f) shows a solution obtained by DFGP for the same task but with a minimum torque constraint applied. }
    \caption{(a) DFGP solution with minimum torque constraint; (b) DFGP solution without minimum torque constraint (See supplemental videos). More details are described in Section~\ref{subsec:kuka_arm}}
	\label{fig:kuka-task2}
	\vspace{-4mm}
\end{figure}

\begin{figure}[!htb]
	\centering
	\includegraphics[scale=0.35]{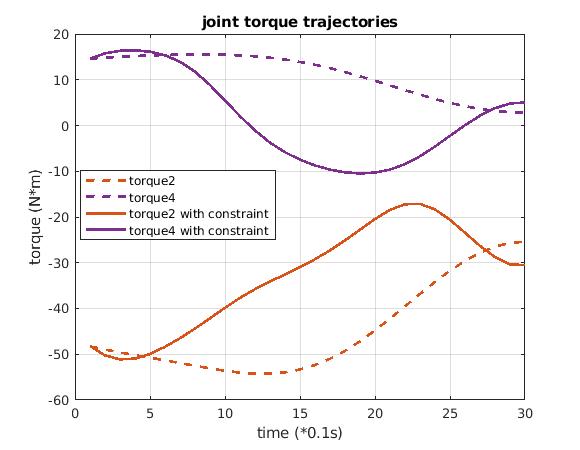}
	\caption{Torque trajectories of Kuka arm experiment. More details are described in Section~\ref{subsec:kuka_arm}.}
	\label{fig:torque}
	\vspace{-5mm}
\end{figure}

The KUKA LBR iiwa~\cite{kuka_webpage} is a lightweight industry robot with seven actuated revolute joints. The task performed in this experiment is to move a block from the floor to its upright position. Fig.~\ref{fig:kukaMinTorque} and Fig.~\ref{fig:kukaNoMinTorque} show solutions obtained by DFGP with and without the minimum torque objective, respectively. In Fig.~\ref{fig:kukaMinTorque}, we observe that the Kuka arm first moves towards the center to reduce the moment arm, and then pushes upwards so that it can bring the block to the goal location with less torque applied when compared to the solution shown in Fig.~\ref{fig:kukaNoMinTorque}. 

We plot the torque trajectories corresponding to the 2\textsuperscript{nd} and 4\textsuperscript{th} joints of the Kuka arm in Fig.~\ref{fig:torque} (the two joints with the highest energy consumption). The solid lines and dashed lines represent the torque trajectories optimized using DFGP with and without the minimum torque constraint. We observe that DFGP produces a more energy-efficient motion plan when we add the minimum torque factor to the graph.

\subsection{Quadruped}
\vspace{2mm}
\begin{figure}[!t]
	\centering
	\begin{subfigure}[b]{0.7\linewidth}
		\includegraphics[width=1.\linewidth]{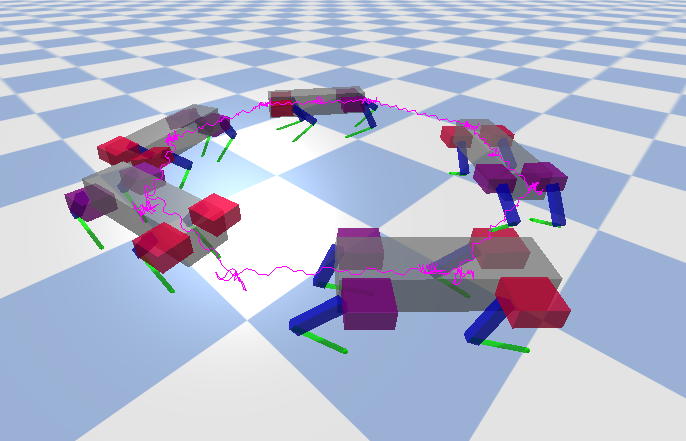}
		\caption{}
		\label{fig:quadruped_task_0}
	\end{subfigure}
	\begin{subfigure}[b]{0.85\linewidth}
		\includegraphics[width=1.\linewidth]{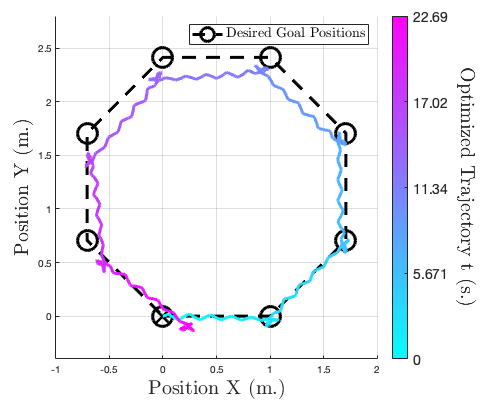}
		\caption{}
		\label{fig:quadruped_task_1}
	\end{subfigure}
	\caption{Trajectory for a quadrupedal robot optimized using DFGP and \emph{executed in an open loop manner}. The dashed line represents the desired CoM trajectory for the given contact schedule. The gradated line represents the actual CoM trajectory when the optimized trajectory is executed in an open-loop fashion. Our method exhibits low error due to its consideration of the full dynamics and kinematics.(See supplemental videos)}
	\label{fig:quadruped_task}
	\vspace{-4mm}
\end{figure}

Motion planning for legged robots is a challenging, high-dimensional optimization problem. The difficulty arises from the contacts, as the addition and removal of contacts with the environment leads to time-varying dynamics constraints. Here we use DFGP to optimize an open-loop trajectory for a 12-DoF quadrupedal robot. We task DFGP with optimizing a trajectory that guides the robot through six navigation waypoints. The waypoints are placed on the vertices of a hexagon with a side length of $1.6m$ (See Fig.~\ref{fig:quadruped_task_1}).

We define a kinodynamic motion planning factor graph and add additional inequality constraints and objectives. Specifically, we incorporate the contact factor defined in Section \ref{ssec:contact_factor} to ensure that the legged robot does not slip. We also apply a minimum torque factor to encourage efficiency and joint limit factors to prevent violation of the robot's joint limits. In this problem, we add a prior over the robot's contact sequence and use GTSAM to optimize for a valid trajectory.

As shown in Fig.~\ref{fig:quadruped_task}, the trajectory generated by our system is highly accurate, leading to little model error \textit{even when executed in an open-loop fashion}. The high tracking accuracy can be explained by our DFG formulation, which models the whole-body dynamics and does not rely on simplified dynamics models.

\section{Discussion}
We introduce dynamics factor graphs (DFGs) as a useful framework for solving various problems in robotics. Our approach treats rigid-body dynamics constraints as factors in a factor graph and optimizes the variables to achieve an objective subjects to dynamics constraints. We also demonstrated how to extend DFGs to support inequality constraints and objective functions. Finally, we introduced DFGP, a motion-planning algorithm that leverages DFGs and trajectory smoothness factors to optimize motion plans for arbitrary robotic systems.

In Section \ref{sec:solving_classical_dynamics_problems}, we demonstrated the application of DFGs to classical dynamics problems, such as forward and inverse dynamics. We saw how our framework intuitively explains various well-known classical dynamics algorithms as DAGs resulting from the solving of the DFG with different elimination orders. We also showed how DFGP can be used to solve practical problems in control and trajectory optimization. We were able to optimize trajectories for three very different robotics systems using a single tool. 

In future work, we hope to perform incremental kinodynamic motion planning in the style of GPMP2~\cite{Mukadam18ijrr} and STEAP~\cite{Mukadam18ar}, which successfully applied incremental inference in factor graphs~\cite{Kaess12ijrr, Dellaert17fnt_fg} to kinematic motion planning problems. Also, it would be exciting to use the factor-graph-based representation of dynamics to perform state estimation for dynamically balanced robots, in the spirit of Hartley \etal~\cite{Hartley18icra_fg_estimation, Hartley18iros_hybrid} and Wisth \etal~\cite{Wisth19ral_legged_fg}.

% In this paper, we represent manipulator dynamics as factor graph and solve for inverse, forward, and hybrid problems. Using factor graphs as a graphical language gives us not only a unified method to solve all types of dynamics problems, but also an insightful visualization of the underlying mathematical formulations. Exploiting different elimination orders of solving the factor graph unlocks powerful tools to illustrate classical algorithms and derive novel algorithms which could be applied to solve certain types of dynamics problems efficiently. 

% As we discussed Section \ref{sec:inverse}, the reported timing results are in no way intended to compete with dedicated dynamics solvers, but rather indicated relative performance.

% In future work we plan to automatically generate code for particular robot topologies, which we hypothesize will match and possibly exceed existing solvers when using non-intuitive but computationally more advantageous elimination orderings. We are also aware that in comparing high performance code controlling for cache effects and memory architecture in general is important, as done by Neuman \etal~\cite{Neuman19iros_benchmarking}.

% \input{sub_tex/4-discussion.tex}

% \section*{ACKNOWLEDGMENT}
% .......................................................................

\clearpage

\bibliographystyle{plain}
\bibliography{references.bib}

\end{document}